\documentclass[letterpaper, 10 pt, conference]{ieeeconf}  

\IEEEoverridecommandlockouts                              

\makeatletter
\let\NAT@parse\undefined
\makeatother

\usepackage{amsmath}
\usepackage{amssymb}
\usepackage{graphics}
\usepackage[tight]{subfigure}
\usepackage[ruled]{algorithm2e}
\usepackage{bm}
\usepackage{algorithmic}
\usepackage{wrapfig}
\usepackage[square, numbers, sort&compress]{natbib}

\usepackage[pagebackref=false,breaklinks=true,letterpaper=true,colorlinks,bookmarks=false,citecolor=blue,linkcolor=blue]{hyperref}
\usepackage{float}
\usepackage{flushend}
\usepackage[capitalize,noabbrev]{cleveref}

\usepackage{booktabs}
\usepackage{tabularx}
\newcolumntype{Y}{>{\centering\arraybackslash}X}

\usepackage{tikz}
\usetikzlibrary{decorations.text,calc,shapes.geometric,shapes.callouts}

\usepackage[export]{adjustbox}

\usepackage[textwidth=1.5cm,textsize=scriptsize,disable]{todonotes}

\newcommand{\todotj}[1]{\todo[fancyline,color=yellow!40]{TJ: #1}\xspace}

\newcommand{\todoty}[1]{\todo[fancyline,color=blue!10]{TY: #1}\xspace}

\newcommand{\todomw}[1]{\todo[fancyline,color=green!60]{MW: #1}\xspace}

\newcommand{\alg}[0]{6-DoFusion\xspace}

\title{\LARGE \bf
6-DoF Stability Field via Diffusion Models
}

\author{
Takuma Yoneda$^{\ast}$\quad
Tianchong Jiang$^{\ast}$ \quad
Gregory Shakhnarovich \quad
Matthew R.\ Walter
\thanks{Takuma Yoneda, Gregory Shakhnarovich, and Matthew R.\ Walter are with the Toyota Technological Institute at Chicago (TTIC), Chicago, IL USA, {\tt\small \{takuma,greg,mwalter\}@ttic.edu}.}
\thanks{Tianchong Jiang is with the University of Chicago, Chicago, IL USA, {\tt\small tianchongj@uchicago.edu}.}
}

\begin{document}

\maketitle
\thispagestyle{empty}
\pagestyle{empty}

\begin{abstract}\todomw{I updated the abstract}
    A core capability for robot manipulation is reasoning over where and how to stably place objects in cluttered environments. Traditionally, robots have relied on object-specific, hand-crafted heuristics in order to perform such reasoning, with limited generalizability beyond a small number of object instances and object interaction patterns\todotj{originally:and classes, but we don't have many classes in the experiments}. Recent approaches instead learn notions of physical interaction, namely motion prediction, but require supervision in the form of labeled object information or come at the cost of high sample complexity, and do not directly reason over stability or object placement. 
    We present \alg, a generative model capable of generating 3D poses of an object that produces a stable configuration of a given scene. Underlying \alg is a diffusion model that incrementally refines a randomly initialized $\textrm{SE}(3)$ pose to generate a sample from a learned, context-dependent distribution over stable poses. We evaluate our model on different object placement and stacking tasks, demonstrating its ability to construct stable scenes that involve novel object classes as well as to improve the accuracy of state-of-the-art 3D pose estimation methods.\todomw{Added reference to pose estimation results}
\end{abstract}
\def\thefootnote{*}\footnotetext{Equal contribution.}\def\thefootnote{\arabic{footnote}}

\section{Introduction}
In order for robots to operate effectively in our homes and workplaces, they must be able to reason over where and how to place objects in our cluttered environments---whether it is to prepare a dining table for a meal or to put dishes away in a cupboard. 
Indeed, arguably the most common use of robot manipulators is for pick-and-place tasks. A common approach to endowing robots with this level of reasoning is to manually define a set of heuristics that attempt to identify valid object placements, for example by detecting empty locations and hard-coding stable orientations for every object~\cite{schmitt2017optimal,xian2017closed,garrett2018ffrob}. These heuristics are often specified in an individual, object- and environment-specific manner, and thus tend to be restricted to relatively few objects\todoty{Are we referring to any specific literature?}\todomw{Added refs}, particularly when the robot needs to reason over placements that involve object interactions (e.g., stacking one mug on top of another). 
The limited ability to generalize to new objects and scenes precludes the use of such heuristics in human environments that are often complex, cluttered, and contain a diverse array of objects~\cite{batra2020rearrangement}.

Data-driven methods provide a promising alternative to the reliance on heuristics to reason over stable object placement~\cite{huang19, zhu21, paxton22}. Indeed, such strategies have the potential to better generalize to novel object and environment instances. However, earlier data-driven methods trade-off hand-engineered heuristics and instead rely on hand-engineered features to represent key properties of inter-object placement~\cite{learning-to-place}. More recent methods instead learn these representations in the context of a model trained to classify the stability of a candidate pose for the objects in the scene, where these candidates may be randomly generated (e.g., by sampling from a uniform distribution) or sampled from a guided distribution~\cite{cheng2021learning}. These stability prediction models typically require access to a large amount of training data in order to be representative of a diverse set of object types and poses, which can be particularly costly when it requires the robot to interact with the environment to collect data. Indeed, approaches that employ rejection sampling can be inefficient, particularly when the support
of the target event is small (e.g., objects requiring careful placement to be stable).

%
%
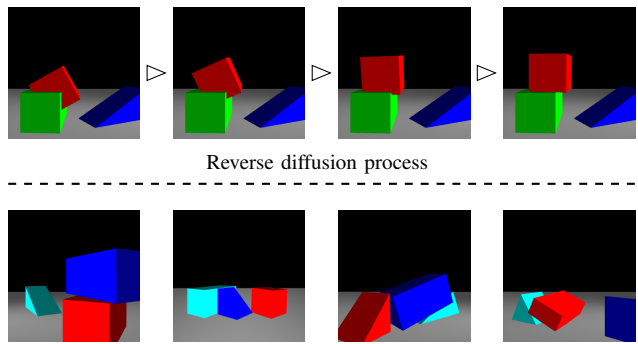
\begin{figure}[t!]
    \centering
    \begin{tikzpicture}[
        triangle/.style={isosceles triangle,draw=black, align=center, minimum size=2mm,align=center,inner sep=0mm}]
        \node[] (gen4) at (0.0,0.0) {\includegraphics[height=1.75cm]{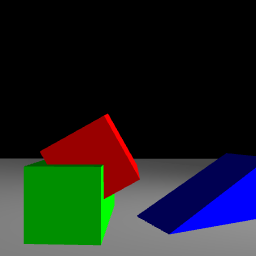}};
        \node[triangle, right = -1pt of gen4] (t3) {};
        \node[right = -1pt of t3] (gen3) {\includegraphics[height=1.75cm]{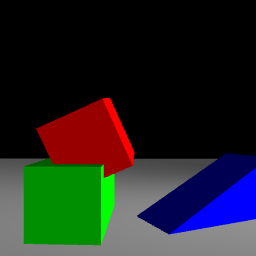}};
        \node[triangle, right = -1pt of gen3] (t2) {};
        \node[right = -1pt of t2] (gen2) {\includegraphics[height=1.75cm]{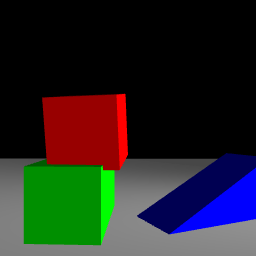}};
        \node[triangle, right = -1pt of gen2] (t1) {};
        \node[right = -1pt of t1] (gen1) {\includegraphics[height=1.75cm]{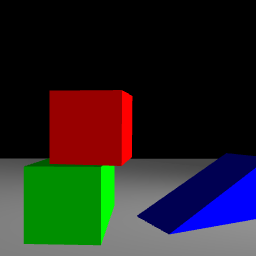}};
        %
        \node[below = 25pt of t2, font=\footnotesize] {Reverse diffusion process};
        \node[below = 10pt of gen4.south west] (div-start) {};
        \node[below = 10pt of gen1.south east] (div-end) {};
        \draw[dashed,thick] (div-start) -- (div-end);
        \node[below = 20pt of gen4] {\includegraphics[height=1.75cm]{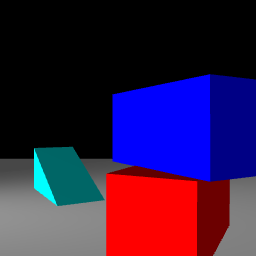}};%
        \node[below = 20pt of gen3] {\includegraphics[height=1.75cm]{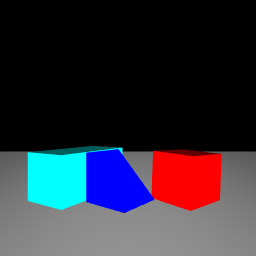}};%
        \node[below = 20pt of gen2] {\includegraphics[height=1.75cm]{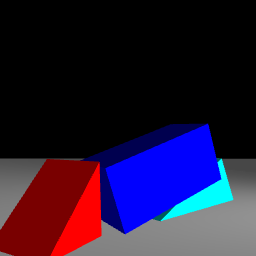}};%
        \node[below = 20pt of gen1] {\includegraphics[height=1.75cm]{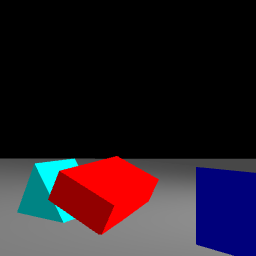}};%
    \end{tikzpicture}%
    \caption{Given (top-left) an initial 3D scene that consists of two objects (a green rectangle and blue wedge), \alg initializes a given object (red cube) in a random pose and (top, left-to-right) employs a reverse diffusion process to generate a (top-right) valid $\textrm{SE}(3)$ pose that results in each object in the scene being stable. As depicted in the bottom row (bottom), \alg is able to produce a diverse array of candidate stable poses.}\label{fig:motivation}
    \vspace{-12pt}
\end{figure}
Instead, we propose \alg (Fig.~\ref{fig:motivation}), a framework that generates poses for a \emph{query} object that stabilizes a given scene (i.e., the \emph{context}) by sampling from a learned context-dependent distribution over stable poses. As a generative model, an advantage of our approach is that it requires access to only positive examples 
of stable scene configurations (object poses), and does not rely upon a separate stable pose classifier nor does it require rejection sampling to produce stable poses. Instead, inspired by the work of \cite{urain2022se3dif}, our \alg framework utilizes a diffusion model~\cite{thermo}, a type of generative model that has proven highly effective at complex generation tasks, for generating samples from a distribution over 6-DoF object poses. Diffusion models are comprised of two primary processes. The first process, referred to as forward process, iteratively adds noise of increasing scale to the input (e.g., the 6-DoF pose of the query object). The second process, known as reverse process, is then trained to iteratively denoise from this noisy input, generating a sample from the desired distribution over stable poses.

Our model reasons over the shape and 6-DoF pose of each object in the workspace, and generates the 6-DoF pose of an additional (query) object, such that it and the existing blocks remain stable. \alg employs a diffusion model to generate this pose as a sample from a learned, context-dependent distribution over stable object poses. We train our model on examples of stable object-object interactions
and evaluate it in scenarios that involve placing and stacking a variety of different object shapes in 3D scenes. We evaluate the quality and diversity of the generated poses and show that \alg is able to successfully place objects to achieve complex object interactions that render an entire scene containing both seen and unseen (novel) objects stable.

\section{Related Work}

Particularly relevant to our work are methods that consider the related problems of object rearrangement~\cite{batra2020rearrangement} and object stacking~\cite{deisenroth11,rusu16,nair18a,li2020towards}. This includes recent neuro-symbolic approaches~\cite{huang19, xu19a, zhu21} to task and motion planning in the context of long-horizon pick and place tasks. These methods are restricted to a set of known objects and do not explicitly reason over the stability of 6-DoF placement pose. 
Perhaps more relevant is the work of \cite{paxton22}, who propose a framework that identifies where and how (in terms of pose) an object can be placed in a stable environment given an RGBD image, such that the resulting pose is both physically stable and consistent with learned semantic relationships. Our method also reasons over how to place an object so as to preserve what is already a stable scene, albeit with regards to the object's full six-DoF pose rather than only its planar rotation as in their work. More significant is that unlike their framework, \alg determines how to place an object so as to make an otherwise unstable environment stable.\todotj{and they cannot make a unstable scene stable?}\todomw{Good catch. Updated}. 

Increasing attention has been paid to the problem of learning physical intuition~\cite{vda, PhysNetNonRigid, li2016fall, lerer16, finn-goodfellow-sergey, npe, yi19, janner2019reasoning, veerapaneni20, piloto22, driess22}. This includes methods that reason over the stability of a given scene as well as those that are concerned with modeling forward dynamics.
Our model differs from this body of work in that it directly generates stable 3D object poses by sampling from a learned distribution, rather than relying on rejection sampling on top of a learned forward prediction model.

In addition to using a notion of stability to guide object manipulation, another line of work explores the benefits of using scene stability as an inductive bias to improve scene understanding. This includes the idea of using learned stability prediction to score candidate scene predictions (e.g., the estimated shape and pose of the objects) \cite{zheng15, indoor-seg-and-support-inf, jia13, scene-understanding-by-reasoning-stability, du18a}, using the intuition that the true scene is stable.

Diffusion models~\cite{thermo} have recently been applied to many problems including image generation, image editing, and text-conditioned image and video generation.\todomw{I added this StructDiffusion reference here. It was in a pargraph by itself below} In the context of robotics, \cite{janner2022diffuser} propose a diffusion-based planning model
that is able to generate a diverse set of feasible trajectories that reach a desired goal.
The ability for diffusion models to generate samples from a potentially multi-modal distribution, a capability that we exploit here, has also been used in the context of other robotics tasks including imitation learning~\cite{pearce2023imitating}, policy learning via offline reinforcement learning (RL)~\cite{diffusion-policy}, and shared autonomy~\cite{yoneda2023noise}. \cite{urain2022se3dif} take advantage of this capability to propose an object-conditioned grasp generation model built on top of a diffusion model, which involves careful considerations of the representation of the $\textrm{SE}(3)$ gripper pose. Very recently, several works \cite{liu2022structdiffusion, simeonov2023rpdiff} propose to use diffusion models for rearrangement task, yet these approaches weigh on generating locally plausible placement poses, rather than considering stability of the entire scene including object interactions as we do.

\section{Method}
\begin{figure*}[!t]
    \centering
    \includegraphics[width=0.75\textwidth]{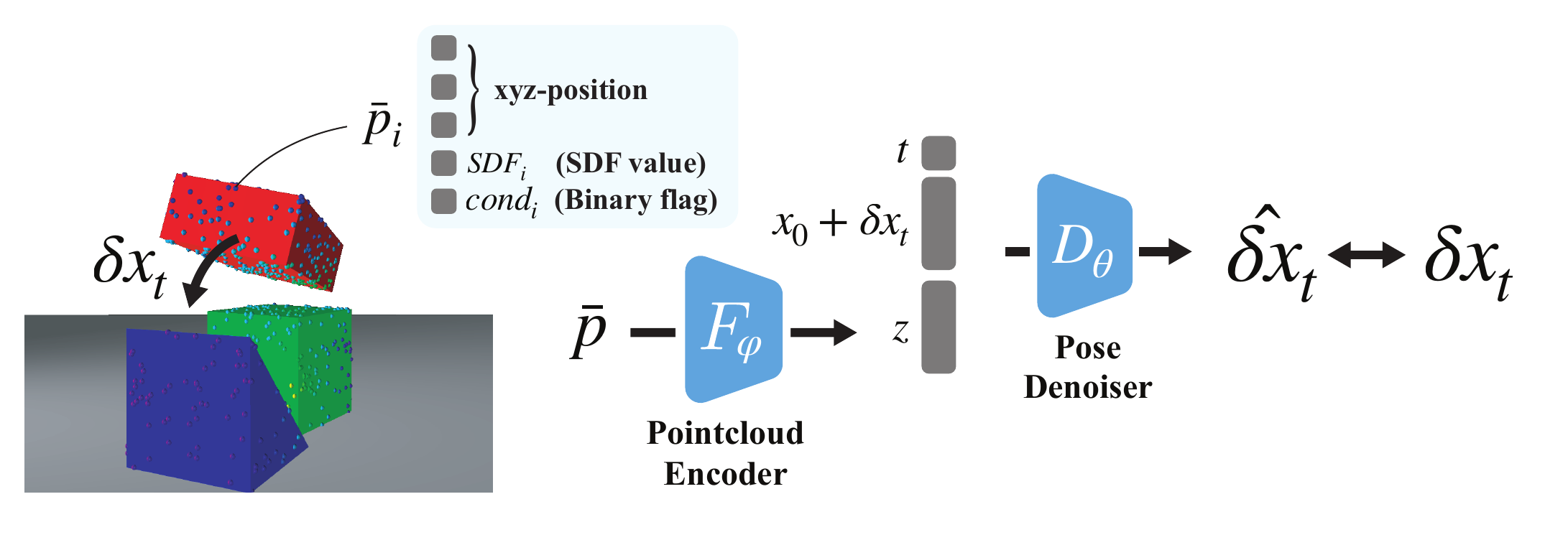}
    \caption{Overview of the presented architecture, where $\bm{x} \in \mathbb{R}^6$ denotes a stable pose sampled from the training dataset, $\delta x_t$ represents the noise sampled from the corresponding noise scale, and $\bar{\bm{p}}$ is a pointcloud with two extra dimensions, a binary flag and SDF value, appended to each point.}
    \label{fig:pipeline}
\end{figure*}
%
%
In this section, after formally define our problem, we describe how to apply diffusion models in $SE(3)$ space, as well as the approach to encode object interactions.
Before introducing our \alg framework (Figure~\ref{fig:pipeline}, Algorithm~\ref{alg:example}), we first provide a mathematical introduction to a general framework of diffusion models~\cite{Karras2022edm}. \todotj{need to change}

\subsection{Background on Diffusion Models}
\label{subsec:diffusion}
Diffusion models~\cite{thermo} are a type of generative model that have proven highly effective for complex generation tasks (e.g., image synthesis), often outperforming their generative adversarial network (GAN)~\cite{saharia2022palette} and variational autoencoder (VAE)~\cite{rombach2022high} counterparts. 
Diffusion models involve two core processes. Modeled as a first-order Markov chain, the \textit{forward diffusion process} involves iteratively adding zero-mean isotropic Gaussian noise of increasing scale $\sigma_t \in \mathbb{R}$ to an initial sample $\bm{x} \sim q(\bm{x})$ drawn from an unknown data distribution. The model is then trained to iteratively denoise a noise-corrupted input through the \textit{reverse diffusion process} that is conditioned on the noise scale $\sigma_t$.
Each step of training a denoising model $D_\theta$ minimizes the following objective
\begin{equation}
\label{eqn:ddpm-objective}
\mathcal{J}_\theta = \mathbb{E}_{\substack{\bm{x}\sim P_\textrm{data} \\ t \sim \textrm{Uniform}[1, T] \\ \bm{\epsilon} \sim \mathcal{N}(\bm{0}, \bm{I})}} \bigg[ \lVert D_\theta (\bm{x} + \sigma_t \bm{\epsilon}, \sigma_t) - \bm{x} \rVert^2\bigg],
\end{equation}
%
%
where $\bm{x}$ is a data sample. In our proposed method, we adopt the denoising diffusion probabilistic model (DDPM)~\cite{ddpm} formulation, where we have a discrete set of noise scales of increasing magnitude $\{\sigma_1, \ldots, \sigma_T\}$, and we sample a noise scale uniformly during training. Once trained, the model generates a sample by iteratively denoising zero-mean isotropic Gaussian noise via the reverse diffusion process. Formally,  this procedure starts with an initial sample \mbox{$\bm{x}_T \sim \mathcal{N}(\bm{0}, \bm{I})$} and involves recursively generating a sample in a previous diffusion timestep, until it reaches $\bm{x}_0$  
\begin{subequations}
    \begin{align}
        \hat{\bm{x}}_0 &= D_\theta(\bm{x}_{t}, \sigma_{t}) \\
        \bm{x}_{t-1} &= \hat{\bm{x}}_0 + \sigma_{t-1} \bm{\epsilon}, ~\bm{\epsilon} \sim \mathcal{N}(\bm{0}, \bm{I}).
    \end{align}
\end{subequations}

\subsection{Problem Formulation}
We consider a setting in which there are $N_\textrm{obj}-1$ objects in the environment with poses $\bm{H}_2, \ldots, \bm{H}_{N_\textrm{obj}} \in \textrm{SE}(3)$ that constitute a \emph{context}.
In this context, we assume that we can place a new \emph{query} object in a pose $\bm{H}_1$ that makes the entire scene stable.
As an example, the context may contain $N_\textrm{obj}-1 = 2$ blocks, one resting on a table and the other elevated above it (e.g., Fig.~\ref{fig:generated-block-poses}, second row)\todoty{This is very hard to understand}.\todomw{I reworded} 
While this configuration is unstable, it is possible to place a new block between the two to support the floating block. In this way, the task is to identify a 6-DoF pose of a query object $\hat{\bm{H}}_1$ such that the full set of objects becomes stable.
We note that there may be many valid poses $\hat{\bm{H}}_1$. We assume access to the object pose $\bm{H}_i \in \textrm{SE}(3)$, pointcloud $\bm{p}_i \in \mathbb{R}^{N_{pts} \times 3}$, and the associated signed distance field (SDF)~\cite{Park_2019_CVPR} $f_i: \mathbb{R}^3 \mapsto \mathbb{R}$ for each object in the workspace ($i \in [2, N_\textrm{objs}]$). We also assume access to the point cloud for the object to be placed and that we can query its SDF for a given point.

\subsection{Diffusion Models in $\textrm{SE}(3)$ Space}

Learning and reasoning over a stable configuration necessitates that we work in $\textrm{SE}(3)$ space.
The $\textrm{SE}(3)$ pose of an object can be represented by a homogeneous matrix $\bm{H} = \left[\begin{smallmatrix} \bm{R} & \bm{t}\\ \bm{0} & 1 \end{smallmatrix}\right] \in \mathbb{R}^{4\times 4}$, where $\bm{R} \in \textrm{SO}(3)$
is a $3 \times 3$ rotation matrix and $\bm{t} \in \mathbb{R}^3$ is the translation vector.

In diffusion models, both the training and generation processes involve iteratively adding Gaussian noise of different scales to the input. 
Applying diffusion model formulation in $\textrm{SE}(3)$ space requires careful consideration of object pose in order to ensure that the transformation matrix remains valid, e.g., to guarantee that $\bm{R} \in \textrm{SO}(3)$. As with \citet{urain2022se3dif}, we represent pose as a six-dimensional vector that consists of the position of the object center $\bm{t} \in \mathbb{R}^3$ in a Cartesian reference frame and an exponential coordinate representation $\bm{e} \in \mathbb{R}^3$ of its orientation. We can add Gaussian noise $\bm{\epsilon}_r \in \mathbb{R}^3$ to an $\textrm{SO}(3)$ orientation as follows,
\begin{equation}
    \tilde{\bm{R}} = \bm{R} \text{Expmap}(\bm{\epsilon}_r),~\bm{\epsilon}_r \sim \mathcal{N}(\bm{0}, \sigma^2 \bm{I}),~ \bm{\epsilon}_r \in \mathbb{R}^3.
\end{equation}

In summary, the procedure for adding noise to a sample $\bm{H}$ from $\textrm{SE}(3)$ space involves (1) sampling noise from an isotropic zero-mean Gaussian $\bm{\epsilon} \sim \mathcal{N}(\bm{0}, \sigma^2 \bm{I})$, where $\bm{\epsilon} \in \mathbb{R}^6$; (2) treating the first three dimensions $\bm{\epsilon}_t = \bm{\epsilon}_{0:3}$ as additive noise on position, and the last three entries $\bm{\epsilon}_r = \bm{\epsilon}_{3:6}$ as noise added to the exponential coordinate; and (3) constructing $\hat{\bm{H}}$ from $\hat{\bm{t}} = \bm{\epsilon}_t$ and $\hat{\bm{R}} = \textrm{Expmap}(\bm{\epsilon}_r)$. Adopting this formulation, we use DDPM~\cite{ddpm} to learn and generate a 6-DoF object pose. 

\subsection{Encoding Object Interactions}
Now that we have established a means of applying diffusion models in $\textrm{SE}(3)$ space, we now describe the process by which we model object interactions. These interactions, which include contacts and object-on-object support relationships are integral to reasoning over the stability of a candidate configuration of objects in the scene. 
In principle, the model can extract such information from the individual pointclouds. 
However, as a form of supplementary information, we include SDF values evaluated on the points as an auxiliary input to the network.   

We define $\bm{p}_{i, n} \in \mathbb{R}^3$ as the coordinates $(x, y, z)$ of point $n$ within the pointcloud of object $o_i$,
and $\bm{H}^\textrm{wo}_i \in \mathbb{R}^{4 \times 4}$ as the transformation matrix that transforms 3D points from the reference frame associated with object $o_i$ to the world frame. Denoting $f_j: \mathbb{R}^3 \mapsto \mathbb{R}$ as a function that returns the SDF for all other objects $j \in [1, N_{\text{objs}}] \setminus \{i\}$, we then compute
\begin{equation}
    s_{i,n} = \min_j f_j(\bm{H}^\textrm{ow}_j \bm{H}^\textrm{wo}_i \bm{p}_{i,n}),
\end{equation}
where $\bm{H}^\textrm{ow}_j = (\bm{H}^\textrm{wo}_j)^{-1}$.
Taking the minimum over SDF values in this fashion can be interpreted as virtually merging all objects with the exception of $o_i$, and computing the SDF value of point $\bm{p}_{i, n} \in \mathbb{R}^3$ with respect to the merged object.
We perform this operation for all objects, including the query object. 
After obtaining the SDF value $s_{i, n} \in \mathbb{R}$ for each point in the pointcloud, we append it to the world-frame coordinate $\bm{p}_{i, n}$
along with a binary flag that distinguishes the set of points that belong to the object being diffused from those that belong to one of the other objects, resulting in a five-dimensional vector.
In the end, we end up with augmented pointcloud $\bar{\bm{p}} \in \mathbb{R}^{N_{\text{objs}} \cdot N_{\text{pts}} \times 5}$.

As depicted in Figure~\ref{fig:pipeline}, we provide the resulting augmented pointcloud as input to the pointcloud encoder 
$F_\varphi: \mathbb{R}^{N_{\text{objs}} \cdot N_{\text{pts}} \times 5} \mapsto Z$
that outputs a vector-valued latent embedding $\bm{z}$ of the augmented pointcloud. This embedding along with a noisy 6-DoF pose of the diffusing object $\bm{x} + \sigma_t \bm{\epsilon} \in \mathbb{R}^6$ and noise scale factor $t$ are then fed to the denoising network $D_\theta$. 
The denoising network predicts the denoised pose $\hat{x}$ and is trained in the same way as a standard DDPM~\cite{ddpm}.

\begin{algorithm}[tb]
    \caption{\alg Training Procedure}\label{alg:example}
    \begin{algorithmic}[1]
       \STATE {\bfseries Input:} pointclouds of objects: $\bm{p}_1, \dots, \bm{p}_n$, \\
       object poses in the workspace: $\bm{H}_2, \dots, \bm{H}_n$
       \WHILE{True}  
       \STATE Sample $t \sim \textrm{Uniform}[0, T]$
       \STATE $\sigma_t \leftarrow \textrm{NoiseSchedule}(t)$
       \STATE $\bm{\epsilon} \sim \mathcal{N}(\bm{0}, \sigma^2_t \bm{I}),~ \bm{\epsilon} \in \mathbb{R}^6$
       \STATE $\tilde{\bm{H}}_i \leftarrow \textrm{AddNoise}(\bm{H}_i, \bm{\epsilon})$
       \FOR{$i, j$ in $1$ to $N_\textrm{objs}$}
       \STATE $\bm{p}_{i,j} \leftarrow \bm{H}_j^{-1}\tilde{\bm{H}}_i \bm{p}_i$ 
       \STATE $\textrm{SDF}_{i,j} \leftarrow \textrm{ComputeSDF}(\bm{p}_{i,j}$, $z_{i})$
       \ENDFOR
       \FOR{$i$ in $1$ to $N_\textrm{objs}$}
       \STATE $\textrm{cond}_i \leftarrow 1~\textbf{if}~i = 1~\textbf{else}~0$
       \STATE $\textrm{SDF}_{i} \leftarrow \min_{j}(\textrm{SDF}_{i,j})~~j \in [1, N_\textrm{objs}] \setminus \{i\}$
       \STATE $\bar{\bm{p}}_{i} \leftarrow \textrm{Concatenate}(\bm{p}_i, \textrm{SDF}_i, \textrm{cond}_i)$
       \STATE $\Psi \leftarrow \textrm{DGCNN}_\varphi(\tilde{\bm{p}}_{i})$
       \STATE $\hat{\bm{\epsilon}} \leftarrow \textrm{Decoder}_\theta (\Psi, \bm{x}_i + \bm{\epsilon}, t)$
       \STATE $\textrm{Optimizer.step}\big(\nabla_{\theta, \varphi} \mathcal{L}(\hat{\bm{\epsilon}}, \bm{\epsilon})\big)$
       \ENDFOR

       \ENDWHILE
    \end{algorithmic}
\end{algorithm}

\section{Experiments}
We evaluate the effectiveness with which our model is able to generate stable poses for seen and unseen (i.e., novel) objects in the context of three task settings: single-block placement, multi-block stacking, and unstructured multi-block placement.
For each domain, 
\alg generates the pose of a single (query) block given the context (i.e., the poses and augmented pointclouds of the other blocks).
We then evaluate the stability of the generated poses by simulating the effects of gravity and object interactions until the objects settle and then measure their displacements.
We note that our task setting, where we aim to generate a $\textrm{SE}(3)$ stable object pose, is quite new that we are not aware of any other work that is directly comparable to ours. For example, \citet{paxton22} address a similar issue, however, they only reason over a planar rotation of the object. \citet{liu2022structdiffusion} adopt an object-centric diffusion model to reason over placement, but their framework expects language conditioning.
As such, we compare against random sampling, followed by forward simulation in some cases as the baseline.

\subsection{Setup}

\subsubsection{Dataset generation}
We consider 3D blocks with seven different shapes. 
These shapes are based on those used by \citet{janner2019reasoning}.
We generate an instance of stable block poses by first randomly sampling block shapes with replacement, dropping them one-by-one from a random $\textrm{SE}(3)$ pose above the ground biased to encourage object interaction, and then use the MuJoCo physics simulator~\cite{mujoco} to simulate the effects of gravity and object interactions. 
We follow this procedure to generate two datasets, one for unstructured block placement and the other for two- and three-object block stacking. We generate $300$K stable block configurations with three blocks, $200$K configurations with two blocks, and $100$K with one block.

\subsubsection{Architecture}
We employ DGCNN~\cite{dgcnn} to encode the augmented pointcloud $\bm{p} \in \mathbf{R}^{N_{\text{objs}} \cdot N_{\text{pts}} \times 5}$ ($F_\varphi$ in Fig.~\ref{fig:pipeline}). DGCNN interleaves the construction of a $k$-nearest-neighbor ($k$-NN) graph on the points and graph convolution. 
For the first layer, we apply $k$-NN graph construction on the four dimensions, i.e., the position and the binary flag for the query object. We adopt the pointcloud classification architecture of \cite{dgcnn}, by removing its classification head. This network gives us a single latent vector $\mathbf{z} \in \mathbb{R}^{512}$. We follow this pointcloud encoder by a three-layer MLP with ReLU activation. The MLP takes in the concatenation of the noisy pose of the query object $\bm{x}$, the diffusion noise scale $t$, and the latent vector $\bm{z}$. The MLP outputs the predicted noise that is then added to the pose of the query object.

\subsubsection{Evaluating a generated pose}

To evaluate the stability of a generated pose, we would ideally want to quantify the difference between the pose and the ``nearest'' stable pose.
However, there is no clear way to identify the ``nearest'' pose in a non-brute-force manner.
Thus, we evaluate the stability of a generated pose by initializing the query object in the MuJoCo simulator at the generated pose, and running forward simulation until every object in the scene settles. 
We then measure the \emph{translational displacement} as the Euclidean distance between the initial and settled object positions, and the \emph{rotational displacement} as the magnitude of the axis-angle representation of the relative rotation. 

We compare our model to a baseline that randomly samples poses above the existing blocks. The baseline samples the orientation of the placing block randomly, and the $(x,y)$ coordinates from a Gaussian centered at the average $(x,y)$ coordinate of the existing blocks. In the block stacking experiment in Section \ref{subsec:single-block-placement}, we set the $(x, y)$ coordinates of the baseline to match the block at the top.

Aside from one model trained on all of the shapes, in order to test the ability of our model to generalize to new shapes, we train seven separate models, each with a different held-out shape.
In evaluation, we split the test dataset into seven subsets, each corresponding to a unique shape. We ensure that the corresponding shape always shows up in every scene of the subset. 
For each test subset, we run a model trained on all shapes (in-distribution, ID) and a model that has not seen the corresponding shape (out-of-distribution, OOD).

Shapes that have longer principle dimensions, such as the long triangle, are more likely to have large translational displacements when settling. As such, we normalize the translational displacements of each block by dividing the raw translational displacement by the diameter of the block. 
%
For translational and rotational displacements, we score each instance using the maximum displacement over all objects (both the context and query objects) and report the median across instances.

\subsection{Single-Block Placement}
\label{subsec:single-block-placement}
As a means of validating that \alg is able to learn to generate reasonable poses,
we first train the model on the single-block placement dataset that includes a single instance of different shapes at stable configurations in the scene.

%
%
    
    

%
\begin{table}[ht]
    \centering
    \caption{Performance on single-block placement}
    \begin{tabularx}{0.95\linewidth}{Xcccc}
    \toprule
     & ID & OOD & Random \\
    \midrule
    Translational Displacement (\%)  & $1.2$ & $2.4$ & $90.7$ \\ 
    
    Rotational Displacement (deg)  & $1.2$ & $1.4$ & $37.2$ \\
    
    \bottomrule
    \end{tabularx}
    \label{tab:single-block-eval}
\end{table}
Table \ref{tab:single-block-eval} summarizes the performance of our model for single-block placement (Fig.~\ref{fig:generated-block-poses}) in terms of the median
translational and rotational displacements over the test set. 
As we see, \alg is able to place the object in stable 3D poses and is also able to generalize to unseen shapes. In contrast, the baseline produces poses that tend to be far less stable, as indicated by the large displacement values.

\subsection{Block Stacking}

Next, we evaluate the ability of our model to place a given query block in a stable pose on top of existing blocks (Fig.~\ref{fig:generated-block-poses}, second and the third rows). We train a single model on a union of the single-block placement and block stacking datasets, where the total number of blocks in a scene ranges from one to three. The evaluation tasks \alg with generating a pose that stacks the given object in a scene that consists of one or two objects.
%
\begin{table}[!ht]
    \centering
    \caption{Success rate (\%) on block stacking (on one block)}
    \begin{tabularx}{0.75\linewidth}{c@{\hspace{1pt}}Xcccc}
    \toprule
    & Block & ID & OOD & Rand.\ Ori.\ \\
    \midrule
        \includegraphics[height=7pt,valign=c]{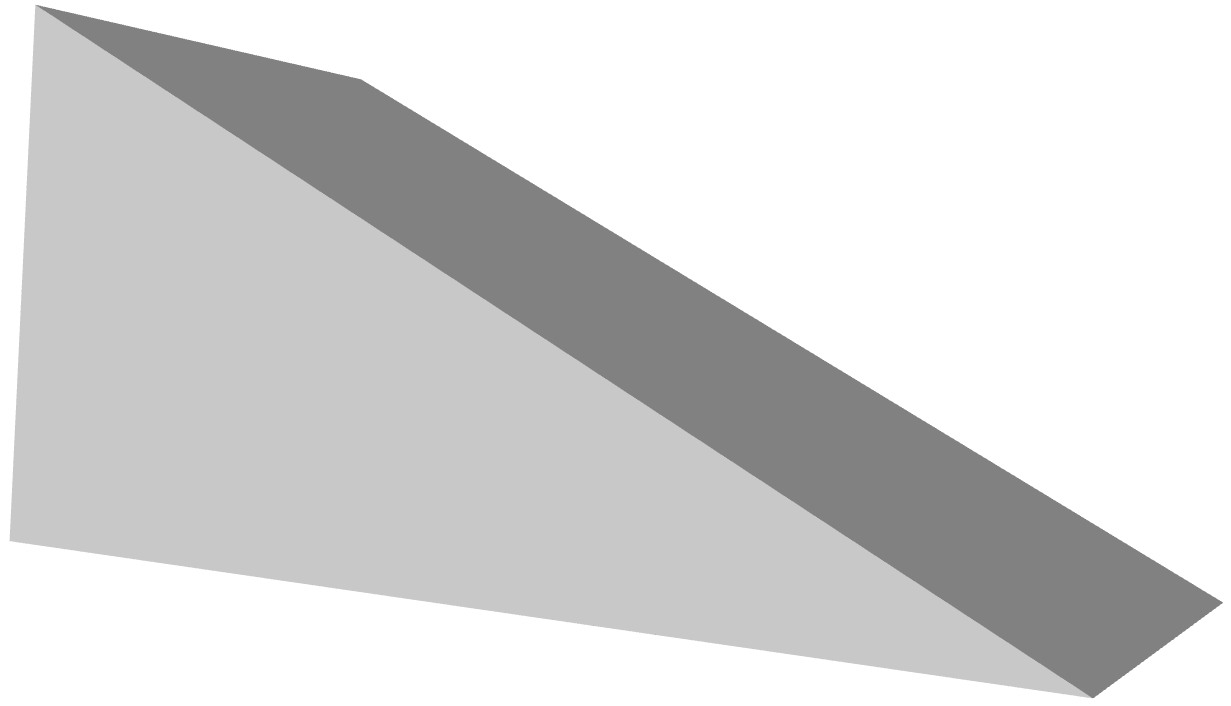} &%
        Tall Triangle   & $81.6$ & $75.6$ & $67.6$ \\
        \includegraphics[height=7pt,valign=c]{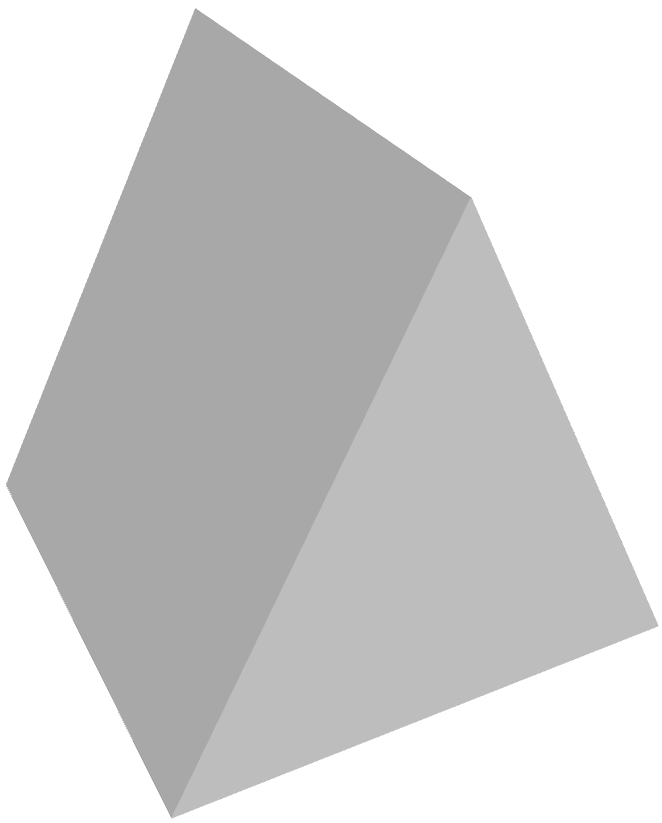} &%
        Middle Triangle & $87.2$ & $83.8$ & $36.2$ \\
        \includegraphics[height=7pt,valign=c]{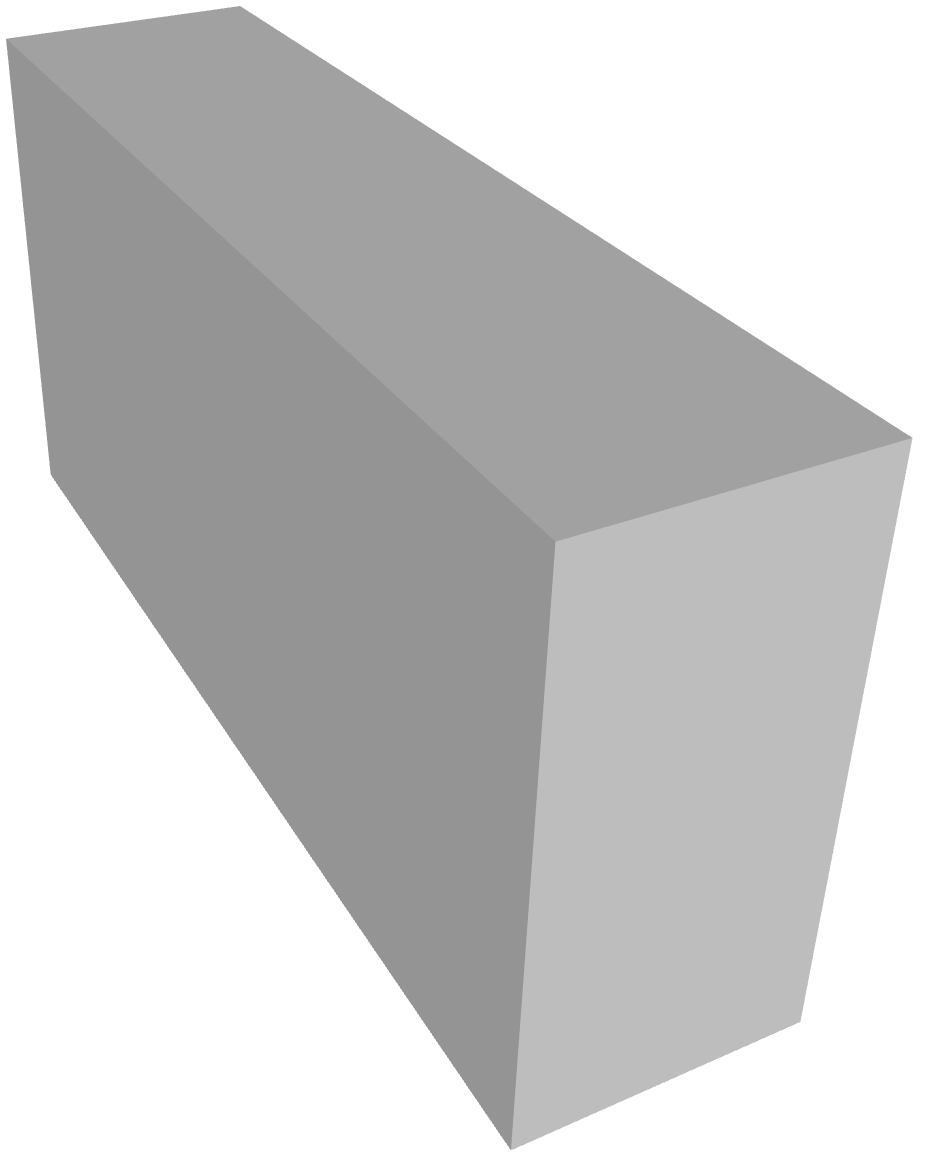} &%
        Half Rectangle  & $87.0$ & $75.6$ & $46.8$ \\
        \includegraphics[height=7pt,valign=c]{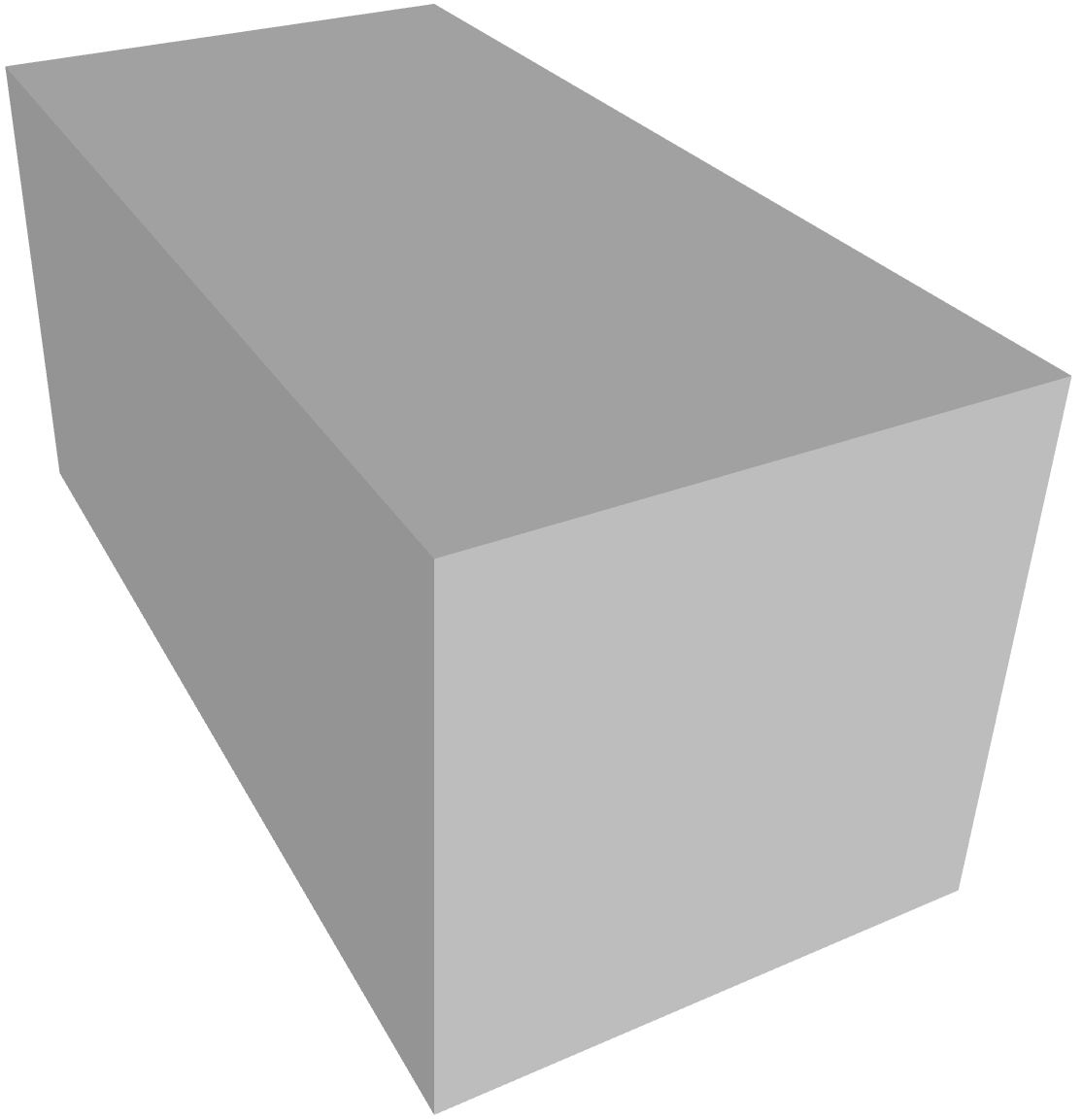} &%
        Rectangle       & $94.6$ & $87.6$ & $51.0$ \\
        \includegraphics[height=7pt,valign=c]{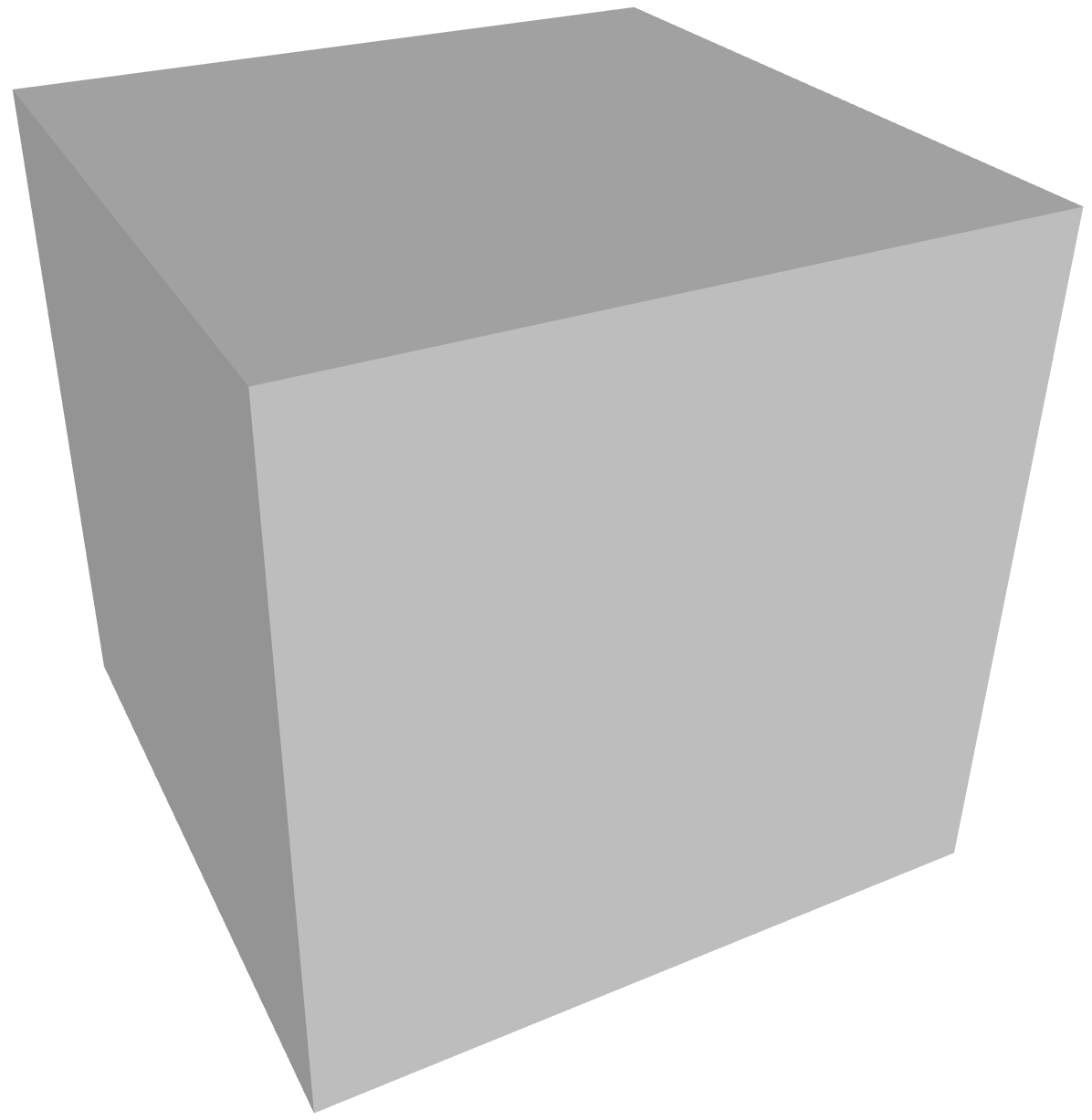} &%
        Cube            & $95.6$ & $89.6$ & $71.6$ \\
        \includegraphics[height=7pt,valign=c]{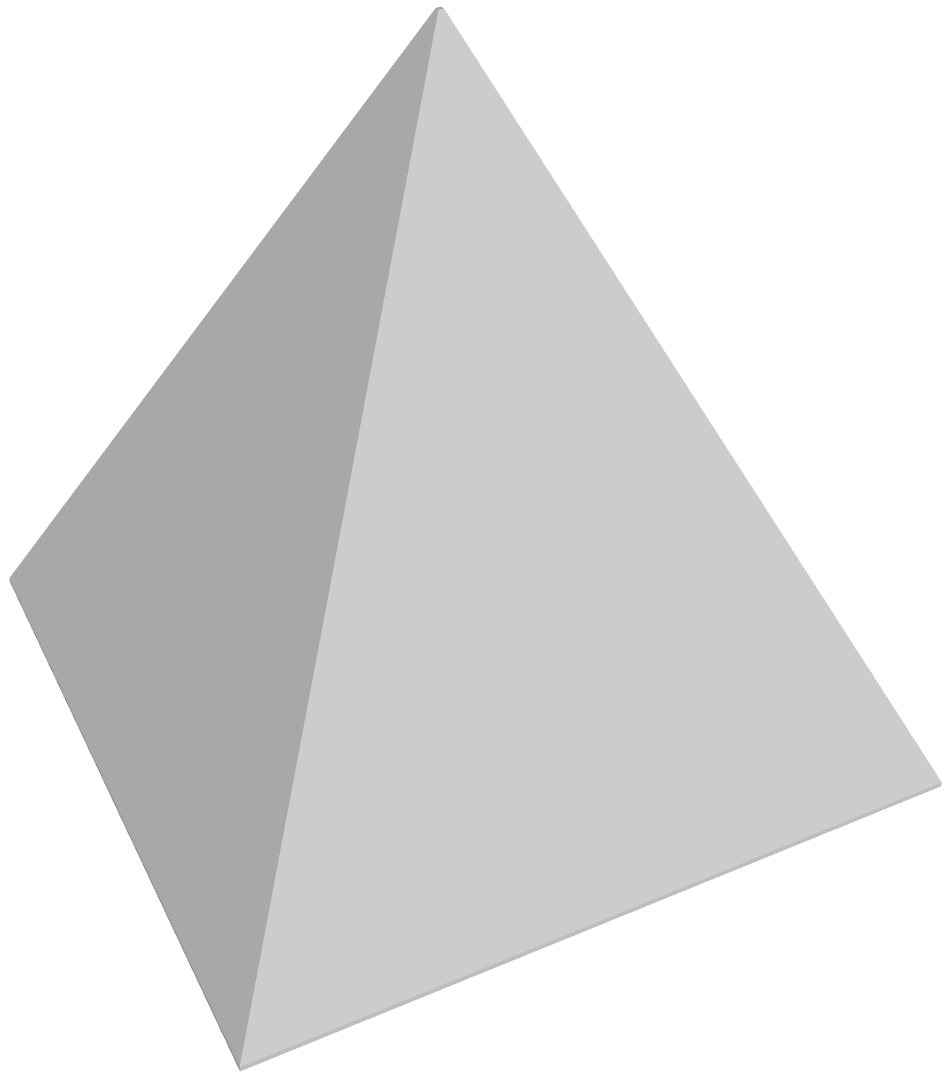} &%
        Tetrahedron     & $80.0$ & $52.2$ & $23.4$ \\
        \includegraphics[height=7pt,valign=c]{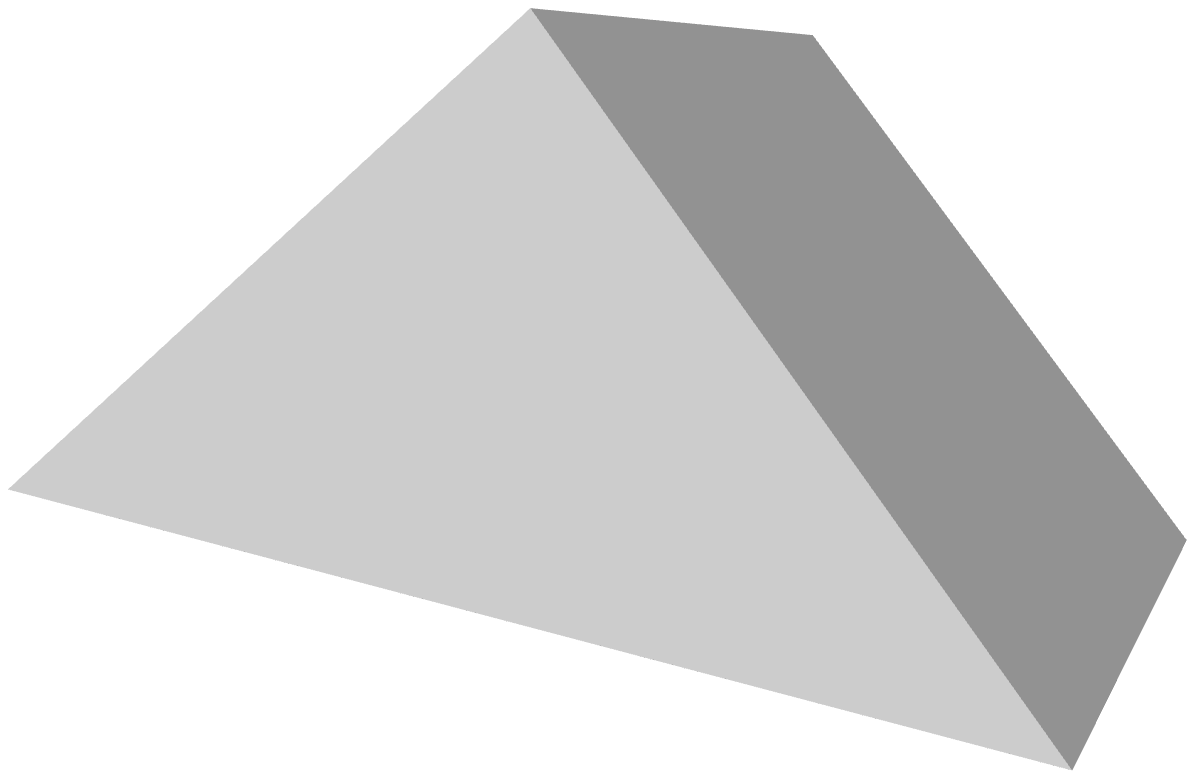} &%
        Hat             & $75.6$ & $68.0$ & $69.8$ \\
    \bottomrule
    \end{tabularx}
    \label{tab:block-stacking-one-block}
\end{table}
We first consider the rate at which \alg successfully stacks a given object in a single-object scene without falling. Table~\ref{tab:block-stacking-one-block} summarizes the in-distribution and out-of-distribution success rates of \alg compared to the baseline, which sets the placing block at the same $(x,y)$ coordinate as the block below, and randomly samples its orientation. We see that \alg performs well on seen and unseen objects. 

%
\begin{table}[!h]
    \centering
    \caption{Success rate (\%) on block stacking (on two blocks)}
    \begin{tabularx}{0.75\linewidth}{c@{\hspace{1pt}}Xcccc}
    \toprule
     & Block & ID & OOD & Rand.\ Ori.\ \\
     \midrule
        \includegraphics[height=7pt,valign=c]{./figs/tall-triangle} &%
        Tall Triangle & $46.6$ & $31.8$ & $18.4$ \\
        \includegraphics[height=7pt,valign=c]{./figs/middle-triangle} &
        Middle Triangle & $53.2$ & $43.8$ & $16.8$ \\
        \includegraphics[height=7pt,valign=c]{./figs/half-rectangle} &
        Half Rectangle & $65.8$ & $54.2$ & $15.4$ \\
        \includegraphics[height=7pt,valign=c]{./figs/rectangle} &
        Rectangle & $80.2$ & $65.0$ & $13.0$ \\
        \includegraphics[height=7pt,valign=c]{./figs/cube} &
        Cube & $90.8$ & $79.6$ & $14.4$ \\
        \includegraphics[height=7pt,valign=c]{./figs/tetrahedron} &
        Tetrahedron & $65.2$ & $26.6$ & $22.4$ \\
        \includegraphics[height=7pt,valign=c]{./figs/hat} &
        Hat & $46.6$ & $18.8$ & $28.2$ \\
    \bottomrule
    \end{tabularx}
    \label{tab:block-stacking-two-blocks}
\end{table}
Table \ref{tab:block-stacking-two-blocks} summarizes the statistics for the scenario in which the initial scene consists of two stacked blocks. This task is significantly more difficult. As the stack is taller, it requires more precise placement (pose) of the block to avoid collapsing the existing stack. It is also indicated by the noticeable decrease in the baseline performance. 
\alg is able to stack blocks successfully in more cases than the baseline even for OOD objects.

\subsection{Conditional Pose Generation}
\begin{table}[!ht]
    \centering
    \caption{Performance of unstructured block placement with two other blocks}\label{tab:three-block-placement-displacement}
    \setlength{\tabcolsep}{2.5pt}
    \begin{tabularx}{\linewidth}{c@{\hspace{1pt}}lYYYY}
        \toprule
        & & \multicolumn{2}{c}{Trans.\ Disp.\ (\%) $\downarrow$} & \multicolumn{2}{c}{Rot.\ Disp.\ (deg) $\downarrow$} \\
        & Block & \alg & Random & \alg & Random\\
        \midrule[0.75pt]
        \multicolumn{6}{c}{\bf In-Distribution (ID)}\\
        \midrule
        \includegraphics[height=7pt,valign=c]{./figs/tall-triangle} & Tall Triangle     &$25.2$ & $\hphantom{0}76.4$  & $26.8$ & $54.6$ \\
        \includegraphics[height=7pt,valign=c]{./figs/middle-triangle} & Middle Triangle &$\hphantom{0}6.8$  & $125.0$ & $\hphantom{0}5.7$  & $50.3$ \\
        \includegraphics[height=7pt,valign=c]{./figs/half-rectangle} & Half Rectangle   &$26.2$ & $\hphantom{0}87.5$  & $22.2$ & $53.5$ \\
        \includegraphics[height=7pt,valign=c]{./figs/rectangle} & Rectangle             &$12.6$ & $\hphantom{0}80.5$  & $10.0$ & $48.9$ \\
        \includegraphics[height=7pt,valign=c]{./figs/cube} & Cube                       &$\hphantom{0}2.3$  & $103.1$ & $\hphantom{0}1.8$  & $43.9$ \\
        \includegraphics[height=7pt,valign=c]{./figs/tetrahedron} & Tetrahedron         &$\hphantom{0}5.5$  & $125.8$ & $\hphantom{0}3.5$  & $45.7$ \\
        \includegraphics[height=7pt,valign=c]{./figs/hat} & Hat                         &$13.4$ & $115.5$ & $12.1$ & $48.1$ \\
        \midrule[0.75pt]
        \multicolumn{6}{c}{\bf Out-of-Distribution (OOD)}\\
        \midrule
        \includegraphics[height=7pt,valign=c]{./figs/tall-triangle} & Tall Triangle     &$17.9$ & $\hphantom{0}78.7$  & $21.0$ & $51.0$ \\
        \includegraphics[height=7pt,valign=c]{./figs/middle-triangle} & Middle Triangle &$\hphantom{0}9.5$  & $125.9$ & $\hphantom{0}7.5$  & $47.3$ \\
        \includegraphics[height=7pt,valign=c]{./figs/half-rectangle} & Half Rectangle   &$32.8$ & $\hphantom{0}86.3$  & $25.7$ & $52.9$ \\
        \includegraphics[height=7pt,valign=c]{./figs/rectangle} & Rectangle             &$40.0$ & $\hphantom{0}76.7$  & $22.0$ & $53.9$ \\
        \includegraphics[height=7pt,valign=c]{./figs/cube} & Cube                       &$\hphantom{0}2.7$  & $105.1$ & $\hphantom{0}2.3$  & $43.9$ \\
        \includegraphics[height=7pt,valign=c]{./figs/tetrahedron} & Tetrahedron         &$\hphantom{0}8.5$  & $132.4$ & $\hphantom{0}5.9$  & $46.6$ \\
        \includegraphics[height=7pt,valign=c]{./figs/hat} & Hat                         &$18.5$ & $110.6$ & $15.2$ & $49.5$ \\
        \bottomrule
    \end{tabularx}
\end{table}

\begin{figure}[t!]
    \centering
    {\setlength{\tabcolsep}{1pt}
    \renewcommand{\arraystretch}{1.5}
    \begin{tabular}{cc|cc|cc}
        \parbox[t]{3mm}{{\rotatebox[origin=c]{90}{{Placement}}}} &%
        \includegraphics[height=43px,valign=c]{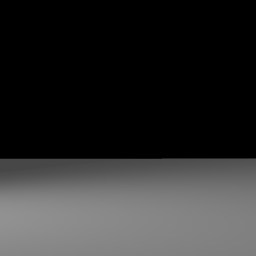} &
        \includegraphics[height=43px,valign=c]{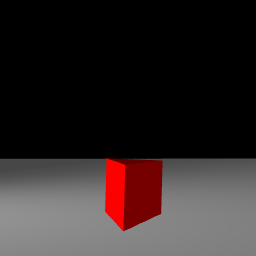} &
        \includegraphics[height=43px,valign=c]{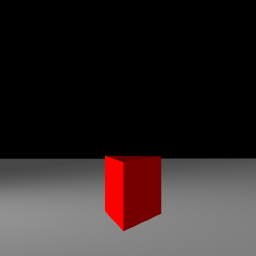} &
        \includegraphics[height=43px,valign=c]{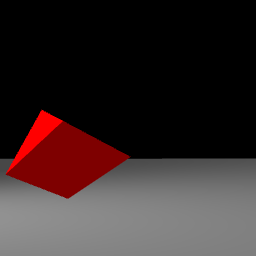} &
        \includegraphics[height =43px,valign=c]{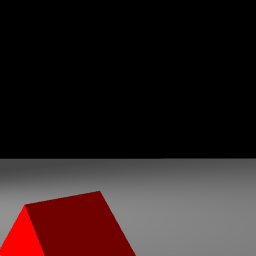}\\[5pt]
        \parbox[t]{3mm}{{\rotatebox[origin=c]{90}{Placement}}} &%
        \includegraphics[height=43px,valign=c]{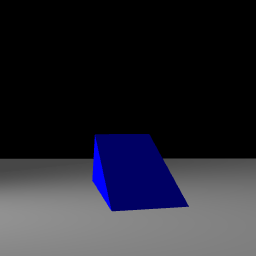} &
        \includegraphics[height=43px,valign=c]{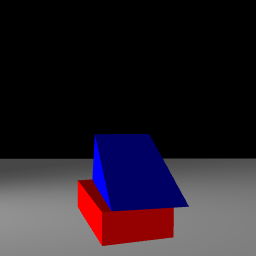} &
        \includegraphics[height=43px,valign=c]{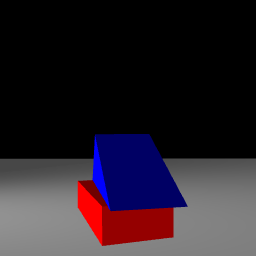} &
        \includegraphics[height=43px,valign=c]{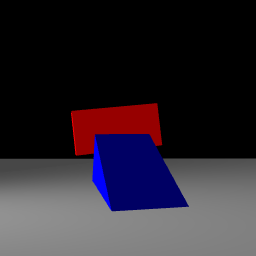} &
        \includegraphics[height=43px,valign=c]{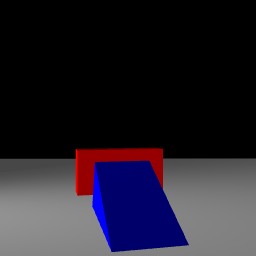}\\[5pt]
        \parbox[t]{3mm}{{\rotatebox[origin=c]{90}{Placement}}} &%
        \includegraphics[height=43px,valign=c]{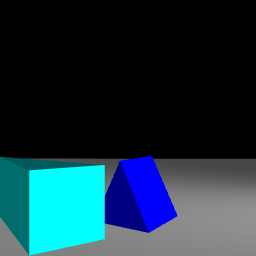} &
        \includegraphics[height=43px,valign=c]{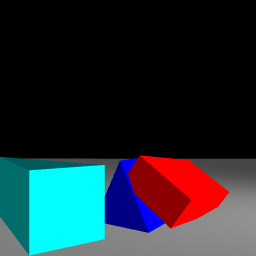} &
        \includegraphics[height=43px,valign=c]{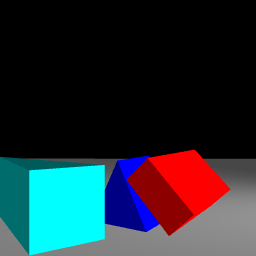} &
        \includegraphics[height=43px,valign=c]{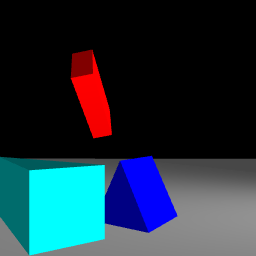} &
        \includegraphics[height=43px,valign=c]{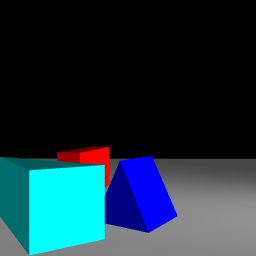}\\[5pt]
        \parbox[t]{3mm}{{\rotatebox[origin=c]{90}{Stacking}}} &%
        \includegraphics[height=43px,valign=c]{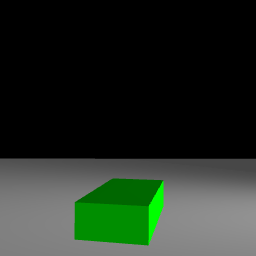} &
        \includegraphics[height=43px,valign=c]{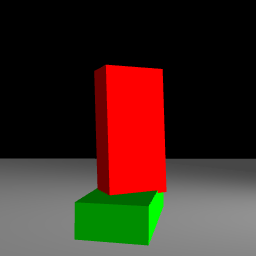} &
        \includegraphics[height=43px,valign=c]{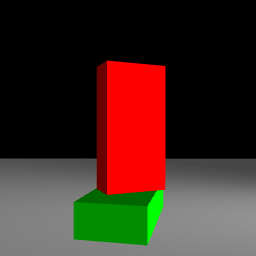} &
        \includegraphics[height=43px,valign=c]{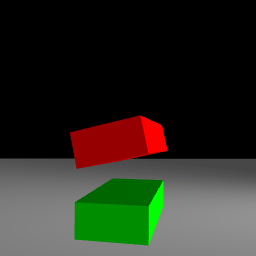} &
        \includegraphics[height=43px,valign=c]{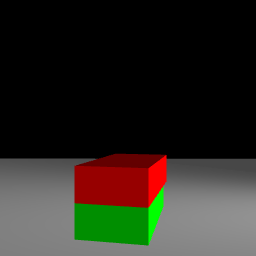}\\[5pt]%
        \parbox[t]{3mm}{{\rotatebox[origin=c]{90}{Stacking}}} &%
        \includegraphics[height=43px,valign=c]{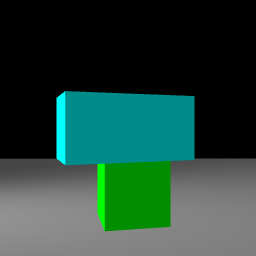} &
        \includegraphics[height=43px,valign=c]{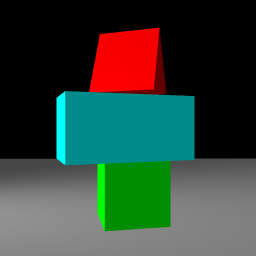} &
        \includegraphics[height=43px,valign=c]{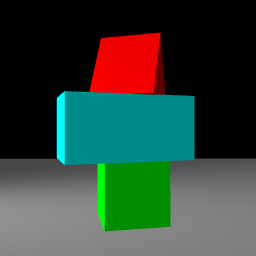} &
        \includegraphics[height=43px,valign=c]{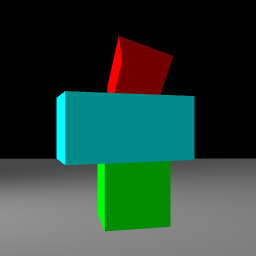} &
        \includegraphics[height=43px,valign=c]{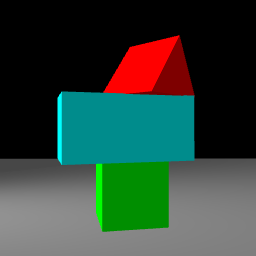}\\[5pt]%
        & Initial & Generated & Settled & Random & Settled
    \end{tabular}}
    \caption{Samples of conditionally generated poses with environments with zero, one, or two blocks and for tasks that involve general placement and object stacking, where each row corresponds to a different scenario. The first column on the left shows the initial scene, while the second column from the left provides a visualization of the pose that \alg generates for the new object in red and the third column shows the pose of the objects after settling. We compare these results to the poses generated by the random baseline (fourth column) and their resulting settled poses (fifth column).}\label{fig:generated-block-poses}
    \vspace{-12pt}
\end{figure}

We consider the more general setting in which the model needs to reason over a placement pose in a cluttered environment (Fig.~\ref{fig:generated-block-poses}, bottom two rows).  
This setting is particularly challenging since the object interactions are far more complex than with block stacking. If a block is leaning on another block, the set of poses is no longer stable if we remove the supporting. In such cases, the model needs to generate poses that are not only stable themselves, but that also support existing blocks in the scene such that the entire configuration is stable.  
We train a single model on the union of the single-block and all unstructured block datasets, where the number of blocks in a scene ranges from one to three.
Table~\ref{tab:three-block-placement-displacement} summarizes the ID and OOD results for scenes with one and two blocks, respectively. Again, we see that \alg produces far more stable poses than the baseline.

%
%
\begin{figure}[t!]
    \centering
    {
    \setlength{\tabcolsep}{1pt}
    \renewcommand{\arraystretch}{1.5}
    \begin{tabular}{ccccccc}
        \parbox[t]{2mm}{{\rotatebox[origin=c]{90}{Initial}}} &%
        \includegraphics[height=1.8cm,valign=c]{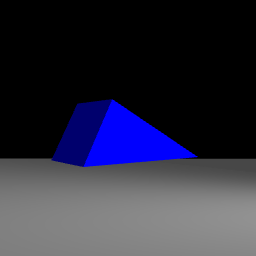} & $\rightarrow$ &%
        \includegraphics[height=1.8cm,valign=c]{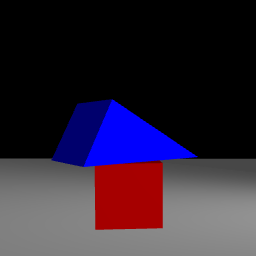} &%
        \includegraphics[height=1.8cm,valign=c]{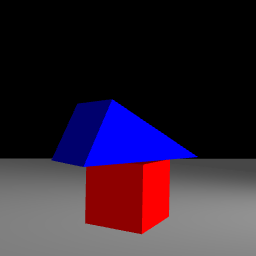} &%
        \includegraphics[height=1.8cm,valign=c]{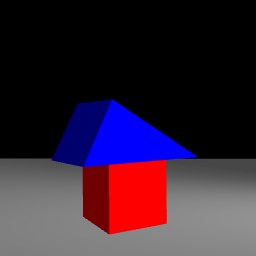} &
        \parbox[t]{2mm}{{\rotatebox[origin=c]{90}{Generated}}}\\[20pt]
        & & & \includegraphics[height=1.8cm,valign=c]{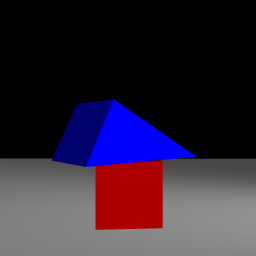} &%
        \includegraphics[height=1.8cm,valign=c]{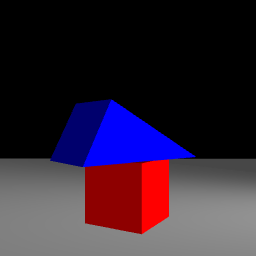} &%
        \includegraphics[height=1.8cm,valign=c]{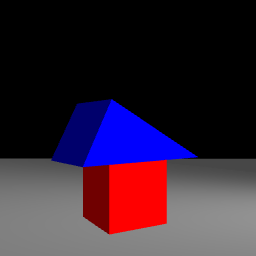}&%
        \parbox[t]{2mm}{{\rotatebox[origin=c]{90}{Settled}}}\\
        
        \midrule
        \parbox[t]{2mm}{{\rotatebox[origin=c]{90}{Initial}}} &%
        \includegraphics[height=1.8cm,valign=c]{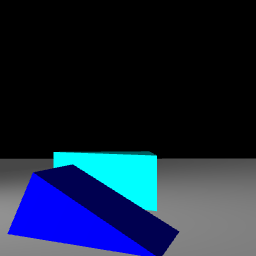} & $\rightarrow$ &%
        \includegraphics[height=1.8cm,valign=c]{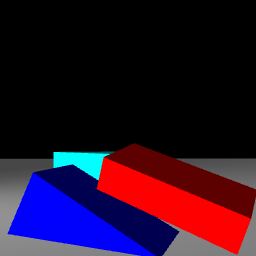} &%
        \includegraphics[height=1.8cm,valign=c]{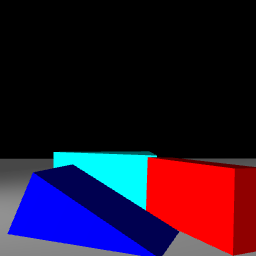} &%
        \includegraphics[height=1.8cm,valign=c]{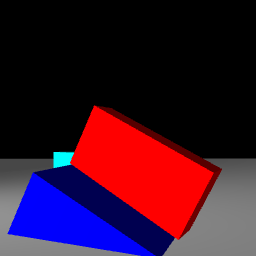} &
        \parbox[t]{2mm}{{\rotatebox[origin=c]{90}{Generated}}}\\[20pt]
        & & & \includegraphics[height=1.8cm,valign=c]{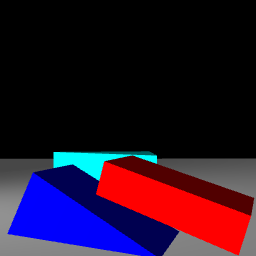} &%
        \includegraphics[height=1.8cm,valign=c]{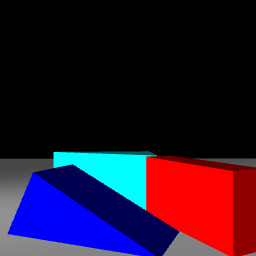} &%
        \includegraphics[height=1.8cm,valign=c]{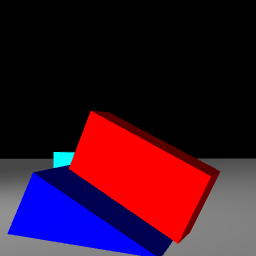}&%
        \parbox[t]{2mm}{{\rotatebox[origin=c]{90}{Settled}}}\\
        \midrule
        \parbox[t]{2mm}{{\rotatebox[origin=c]{90}{Initial}}} &%
        \includegraphics[height=1.8cm,valign=c]{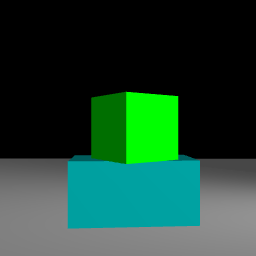} & $\rightarrow$ &%
        \includegraphics[height=1.8cm,valign=c]{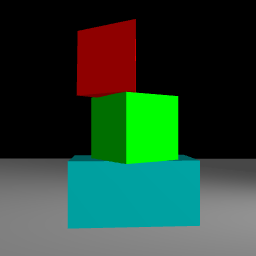} &%
        \includegraphics[height=1.8cm,valign=c]{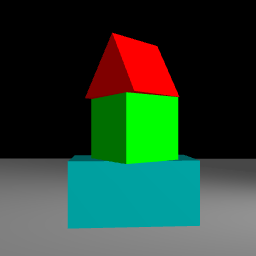} &%
        \includegraphics[height=1.8cm,valign=c]{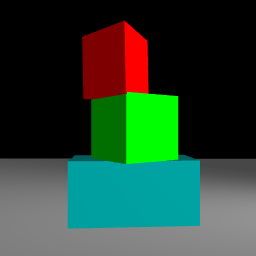} &
        \parbox[t]{2mm}{{\rotatebox[origin=c]{90}{Generated}}}\\[20pt]
        & & & \includegraphics[height=1.8cm,valign=c]{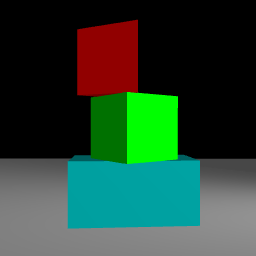} &%
        \includegraphics[height=1.8cm,valign=c]{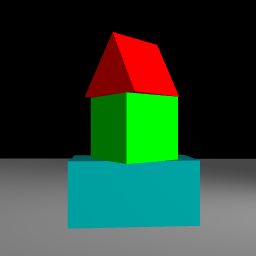} &%
        \includegraphics[height=1.8cm,valign=c]{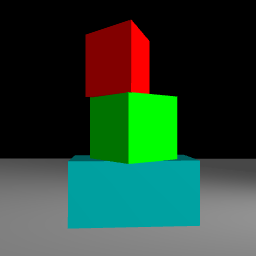}&%
        \parbox[t]{2mm}{{\rotatebox[origin=c]{90}{Settled}}}
    \end{tabular}}

    \caption{Examples demonstrating the diversity with which our algorithm generates poses of a new object (shown in red) for three different initial scene contexts. Each of the three columns on the right visualizes (top row) different poses that \alg generates for the new object along with the (bottom row) settled poses. Note that in the top scenario, \alg stabilizes the blue triangle by placing the given object underneath it.} \label{fig:generation-diversity}
\end{figure}

\subsection{Ablation Studies}

Next, we ablate the different components of our framework in order to better understand their contribution. In particular, we consider the effect of removing SDF information as an explicit indication of inter-object geometry, as well as the effect of removing explicit pose of the placing object from the input to the model.

We consider the task of placing a block in an unstructured scene that contains two blocks, 
and compute the translational and rotational displacements, as well as the success rates.

%
\begin{table}[!ht]
    \centering
    \caption{Ablation study}
    \label{tab:ablation}
    {\setlength{\tabcolsep}{0pt}
    \begin{tabularx}{1.0\linewidth}{lYYYYYY}
        \toprule
        & \multicolumn{3}{c}{In-distribution} & \multicolumn{3}{c}{Out-of-distribution}\\
        Block & Full & w/o SDF & w/o pos & Full & w/o SDF & w/o pos \\
        \midrule
        Trans. Displ. (\%) & $13.1$ & $18.5$ & $70.0$ & $18.6$ & $27.7$ & $82.7$\\
        Rot. Displ. (deg) & $11.7$ & $15.6$ & $74.0$ & $14.3$ & $25.5$ & $80.5$ \\
        \bottomrule
    \end{tabularx}}
\end{table}
Table~\ref{tab:ablation} summarizes the result of these ablations. This suggests that the model gains meaningful information from both the SDF and the explicit pose of the placed object.

\section{Conclusion}
We presented \alg, a diffusion-based model that generates 6-DoF object poses that result in stable multi-object scenes. We evaluated our model on multi-object placement and stacking tasks, and demonstrated that it is able to reason over the full 6-DoF pose of novel objects placed in complex, unstructured scenes. A limitation of \alg is that it assumes knowledge of the pose and SDF of relevant objects. Future work includes updating the model to reason over noisy estimates of object poses and pointclouds, such as those that can be inferred from RGBD images of the scene.





\flushend



\bibliographystyle{IEEEtranN}
\bibliography{references}
\flushend

\end{document}